\pdfoutput=1
\documentclass[11pt]{article}

\usepackage{acl2023}

\usepackage{times}
\usepackage{latexsym}
\usepackage{amssymb}
\usepackage[T1]{fontenc}
\usepackage[utf8]{inputenc}
\usepackage{microtype}
\usepackage{inconsolata}
\usepackage{graphicx}
\usepackage{longtable}
\usepackage{tabularx}
\usepackage{booktabs}
\usepackage{multicol}
\usepackage{dblfloatfix}
\usepackage{fvextra}
\usepackage{todonotes}
\usepackage{tikz}

\usetikzlibrary{babel,positioning,fit,matrix,backgrounds,arrows.meta,shapes.geometric,shadows,calc,patterns.meta}

\usepackage[subtle]{savetrees}

\usetikzlibrary{patterns.meta}
\usepackage[most]{tcolorbox}

\DeclareUnicodeCharacter{26A0}{} % ⚠ 
\DeclareUnicodeCharacter{FE0F}{}                      % ignore 
\DeclareUnicodeCharacter{2265}{$\geq$}  

\usepackage{fvextra}   % extends fancyvrb; works on arXiv
\usepackage{mdframed}  % lightweight framed boxes

\mdfdefinestyle{pa@box}{%
  linewidth=0.6pt,
  roundcorner=3pt,
  skipabove=\baselineskip,
  skipbelow=\baselineskip,
  innerleftmargin=8pt,
  innerrightmargin=8pt,
  innertopmargin=6pt,
  innerbottommargin=6pt
}

\newcommand{\easySymbol}{\tikz[baseline=-0.2ex]\draw (0,0) rectangle (0.8em,0.8em);}
\newcommand{\hardSymbol}{
  \tikz[baseline=-0.2ex]
    \filldraw[
      pattern={Dots[distance=3pt, radius=0.4pt]},
      pattern color=black
    ] (0,0) rectangle (0.8em,0.8em);
}

\title{
Reasoning Core: A Scalable Procedural Data Generation Suite\\
for Symbolic Pre-training and Post-Training
}

\newlength{\myMheight}
\newcommand{\github}{\includegraphics[height=\myMheight]{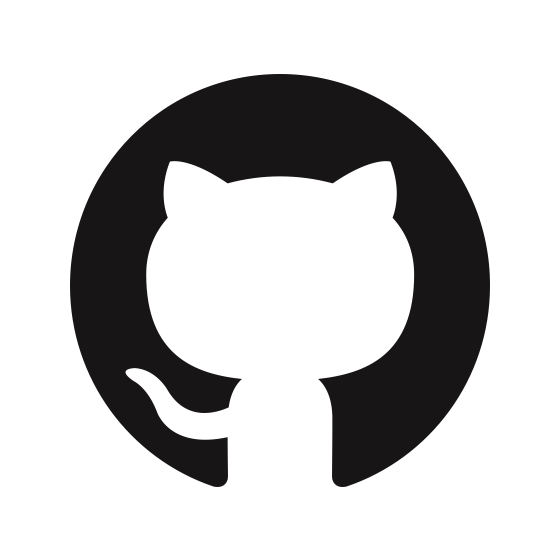}}
\newcommand{\hf}{\includegraphics[height=\myMheight]{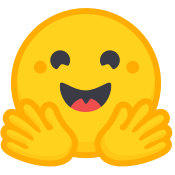}}
\newcommand{\pitl}{\includegraphics[height=\myMheight]{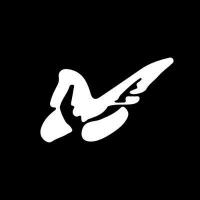}}

\author{Valentin Lacombe \and Valentin Quesnel \and Damien Sileo \\
Univ. Lille, Inria, CNRS, Centrale Lille, UMR 9189 - CRIStAL, F-59000 Lille, France \\
\texttt{damien.sileo@inria.fr}}

\begin{document}
\maketitle

\begin{abstract}

Training on verifiable symbolic data is a promising way to expand the reasoning
frontier of language models beyond what standard pre-training corpora provide.
Yet existing procedural generators often rely on fixed puzzles or templates and do not deliver the distributional breadth needed at scale. We introduce Reasoning Core, a scalable suite that procedurally generates verifiable symbolic reasoning data across core formal domains: PDDL planning over randomized domains, first-order logic with equality, context-free grammar parsing and generation, causal reasoning over random Bayesian networks, and systems of equations. Each task is paired with an external solver for rigorous verification and admits continuous difficulty control for curriculum design. Examples can optionally include solver-derived reasoning traces, enabling supervised training from the earliest pre-training stages, and the same interface provides verifiable reward functions for reinforcement learning. Our experiments show that mixing Reasoning Core data into pre-training improves downstream reasoning while preserving, or slightly improving, language modeling quality. Zero-shot evaluations confirm these tasks challenge frontier models such as GPT-5.
The code and data are publicly available under the MIT license\footnote{
  \begin{tabular}{@{}ll@{}}
   % \yt & \href{https://youtu.be/M37pbdJo5XA}{\texttt{youtu.be/M37pbdJo5XA}} \\
    \github & \href{https://github.com/sileod/reasoning_core}{\texttt{github.com/sileod/reasoning\_core}} \\
    \hf & \href{https://hf.co/collections/reasoning-core/datasets}{\texttt{hf.co/collections/reasoning-core/datasets}}\\
    \pitl & \href{https://app.primeintellect.ai/dashboard/environments/sileod/reasoning-core-env}{\texttt{environments/sileod/reasoning-core-env}}
  \end{tabular}
}.
\end{abstract}

\section{Introduction}

As language models scale, there is growing interest in training on verifiable symbolic data to build reasoning capabilities, rather than relying solely on web text \cite{Wu2022InsightsIP,liu2023zero,Akter2025FrontLoadingRT, jiang2026proceduralpretrainingwarminglanguage}.
Evidence suggests that post-training methods like Reinforcement Learning with Verifiable Rewards \cite{r1, tulu3, o1} (RLVR) often amplify strategies already latent from pre-training \cite{Zhang2025OnTI, yue2025doesreinforcementlearningreally, yuan2025fxgxfgxllms, matsutani2025rlsqueezessftexpands}. Going beyond these capabilities likely requires exposure to broader problem distributions. This scale demands either compute-intensive prolonged RL~\cite{liu2025prorl} or a complementary approach like \emph{symbolic pre-training}: injecting algorithmically generated formal tasks into early training data to build foundational reasoning primitives that post-training can exploit.

Procedural generation is a natural instrument for producing such data, but its effectiveness depends critically on distributional breadth. Narrow templates and fixed puzzles can probe whether a model already possesses a capability, but may not provide the diversity needed to \emph{instill} new reasoning primitives at scale. Training on a single PDDL planning domain such as BlocksWorld does not generalize to minor variations~\cite{valmeekam2023planbench, khandelwal2024pddlfuse}, and Dyck-language pre-training~\cite{hu-etal-2025-circuits, jiang2026proceduralpretrainingwarminglanguage} covers only a narrow slice of hierarchical structure. Existing post-training suites like Reasoning Gym~\cite{stojanovski2025reasoning} provide excellent signals to train reasoning models through RLVR, but they prioritize the number of tasks over distributional generality within each task and do not target pre-training-scale data production.

To address this gap, we introduce \textbf{Reasoning Core}, a scalable suite of procedural generators designed to produce verifiable symbolic data with high distributional generality for both pre-training and post-training. Reasoning Core offers fewer task generators than broad-coverage libraries but targets more foundational and expressive domains: PDDL planning over randomly generated domains (not fixed scenarios), full first-order logic with equality, context-free grammar parsing and generation with arbitrary grammars, causal reasoning over random Bayesian networks, and symbolic equation solving. Each generator is paired with an external solver for rigorous verification and a continuous difficulty control for curriculum design, yielding an effectively unbounded supply of novel instances. The same interface exposes both training-ready examples and a verifiable reward function, so the generated data can serve pre-training, instruction-tuning, and future reinforcement learning.
Our contributions are:
\begin{itemize}\setlength\itemsep{0pt}
\item[-] A scalable generation suite covering core formal domains (PDDL planning, first-order logic, context-free grammars, causal reasoning, equation solving) with high distributional generality and external solver verification.
\item[-] \texttt{gramforge}, a grammar framework with topological control and context-sensitive derivation for producing structured symbolic data.
\item[-] Pre-training experiments showing that mixing Reasoning Core data improves reasoning while preserving, or even slightly improving, language modeling quality.
\item[-] Publicly released datasets (5B pre-training tokens, 2B post-training tokens) and code under the MIT license.
\end{itemize}

\section{Related Work}

\paragraph{LLM-generated and application-oriented synthetic data} LLMs are frequently used to generate synthetic data for targeted domains. Notable examples include MathGenie~\cite{lu-etal-2024-mathgenie}, which employs question back-translation for mathematical reasoning, and DeepSeek-Prover~\cite{Xin2024DeepSeekProverATA} for theorem proving. Related methods address coding~\cite{xu2025kodcode} and instruction tuning~\cite{kaur2024instructskillmixpowerfulpipelinellm}. While effective for post-training, these LLM-based techniques incur high inference costs and offer limited distributional breadth. In contrast, our purely procedural suite provides scalable, verifiable symbolic data with high generality, and is well-suited for pre-training.

\paragraph{Procedurally Generated Environments}
To overcome static dataset limitations, procedural content generation (PCG) has become popular for creating dynamic and scalable evaluation environments~\cite{cobbe2020leveraging, risi2020increasing, liu2025synlogic, liu2026scaling}. Reasoning Gym~\cite{stojanovski2025reasoning} is a state-of-the-art example, offering a large suite of algorithmically verifiable tasks with parametric difficulty. Other work has focused on procedurally generating specific types of challenges, such as games~\cite{balrog, seely2025sudokubenchevaluatingcreativereasoning}, logic puzzles~\cite{lin2025zebralogicscalinglimitsllms, lee2025fol}, or visual reasoning problems~\cite{chollet2025arc, wu2025nesygeo}. Reasoning Core focuses on the full generality of fundamental symbolic domains like planning and formal logic, deliberately using fewer tasks with far greater distributional generality. For example, RG offers individual tasks such as Sokoban or Hanoi towers, which are specific PDDL planning domains, whereas RC samples randomized PDDL domains that cover the full class of STRIPS problems.
For logic, RG hardcodes syllogisms, while RC employs formal semantics via a state-of-the-art first-order logic grammar~\cite{sileo-2024-scaling}.
We also provide metadata and algorithmic chain-of-thought traces when possible. We expand on these differences in Section~\ref{sec:rc_features} and summarize them in Table~\ref{tab:comparison}.

\begin{table}[t]
\centering\small
\begin{tabularx}{\columnwidth}{@{}lcc@{}}
\toprule
\textbf{Feature} & \textbf{RC} & \textbf{RG} \\
\midrule
Task families & 28 & 100+ \\
Distributional generality & High & Medium/Templates \\
External solver verification & $\checkmark$ & Partial \\
Chain-of-thought traces & $\checkmark$ & -- \\
Pre-training support & $\checkmark$ & -- \\
Continuous difficulty knob & $\checkmark$ & Param. \\
Grammar-based generation & $\checkmark$ & -- \\
\bottomrule
\end{tabularx}
\caption{Comparison of Reasoning Core (RC) and Reasoning Gym~\cite{stojanovski2025reasoning} (RG). The two suites are complementary: RG provides broad coverage of puzzle and game tasks, while RC prioritizes distributional generality over core formal domains (e.g., sampling arbitrary STRIPS problems rather than fixed instances).}
\label{tab:comparison}
\end{table}

\begin{figure*}[t]
\centering
\newcommand{\codefont}{\ttfamily\fontsize{6.5}{8}\selectfont\color{black!85}}
\begin{tikzpicture}[
font=\sffamily,
every node/.style={outer sep=0pt},
>={Latex},
apibox/.style={
draw=black!35, fill=black!3, rounded corners=6pt,
inner sep=6pt, align=left, text width=0.96\textwidth
},
panel/.style 2 args={
draw=#1!55, fill=#1!4, rounded corners=6pt,
inner sep=6pt, align=left, text width=#2
},
pill/.style={
fill=black!6, draw=black!15, rounded corners=8pt,
inner sep=5pt, align=center, text width=0.96\textwidth
},
arrow/.style={->, thick, draw=black!45},
]
% --- 1. API Box (Top) ---
\node[apibox] (api) at (0,0) {\textbf{Task interface} (\texttt{
28 task names in list\_tasks()})
\vspace{2pt}\par
{\ttfamily\small
t = get\_task(\textit{name}) \quad|\quad
ex = t.generate\_example(level=\textit{k}) \quad|\quad
r = t.score\_answer(\textit{ans}, ex)
}};
% --- 2. Task Panels (Aligned left & right to prevent straying) ---
\node[panel={violet}{0.465\textwidth}, anchor=north west] (p1) at ([yshift=-5mm]api.south west) {%
\textbf{\textsf{arithmetics}} \hfill {\footnotesize\color{black!55}verifiable numeric answer}\par
\vspace{2pt}
{\codefont
\textcolor{violet!70!black}{\# generate an example at difficulty level 2}\par
t = get\_task('arithmetics')\par
ex = t.generate\_example(level=2)\par\vspace{2pt}
\textcolor{violet!70!black}{\# prompt (for SFT / RL rollouts)}\par
Evaluate (3+4.5)*min(8,12)-2**2.\par
Answer with only a number.\par\vspace{2pt}
\textcolor{violet!70!black}{\# answer + optional trace (for training)}\par
ex.answer = "56"\par
ex.metadata.cot = "3+4.5=7.5; min(8,12)=8; 7.5*8=60; 60-4=56"\par
}};
\node[panel={teal}{0.465\textwidth}, anchor=north east] (p2) at ([yshift=-5mm]api.south east) {%
\textbf{\textsf{logic\_nli}} \hfill {\footnotesize\color{black!55}verified by external prover}\par
\vspace{2pt}
{\codefont
t = get\_task('logic\_nli')\par
ex = t.generate\_example(level=3)\par\vspace{2pt}
\textcolor{teal!70!black}{\# prompt (natural language)}\par
Premise: Everyone kind is happy.\par
If happy, they are wise. Mary is kind.\par
Hypothesis: Mary is wise.\par
... .\par\vspace{2pt}
\textcolor{teal!70!black}{\# t.prompt is exposed -- reformulate freely}\par
t.prompt = my\_custom\_prompt\_fn\par
ex = t.generate\_example(level=3) \textcolor{teal!70!black}{\# new format}\par
}};
% --- 3. Difficulty bar ---
\node[pill, anchor=north] (diff) at ([yshift=-3mm]api.south |- p1.south) {%
\textbf{Continuous difficulty control} (single knob per task)\\[3pt]
\begin{tikzpicture}[baseline=0]
% Outer container (white background) perfectly snapped to content
\draw[draw=black!25, fill=white, line width=0.8pt, rounded corners=4pt]
(0,-0.35) rectangle (13.2,0.55);
% Main axis line
\draw[->, thick, draw=black!40] (0.2,0.0) -- (12.6,0.0);
% Pre-training block (Levels 0-2)
\fill[blue!5, rounded corners=2pt] (0.8, 0.05) rectangle (6.2, 0.35);
\draw[|-|, draw=blue!50, thick] (1.0, 0.2) -- (6.0, 0.2)
node[midway, above, inner sep=2pt, text=blue!80!black, font=\sffamily\fontsize{6.5}{7}\selectfont] {Pre-training \& SFT (Base Models)};
% Post-training block (Levels 3-5)
\fill[red!5, rounded corners=2pt] (7.2, 0.05) rectangle (12.2, 0.35);
\draw[|-|, draw=red!50, thick] (7.4, 0.2) -- (12.0, 0.2)
node[midway, above, inner sep=2pt, text=red!80!black, font=\sffamily\fontsize{6.5}{7}\selectfont] {Post-training \& RLVR (Frontier Models)};

\foreach \x/\lab in {1.0/0, 3.5/1, 6.0/2, 7.4/3, 9.7/4, 12.0/5} {
\draw[draw=black!40, thick] (\x,-0.05) -- (\x,0.05);
\node[font=\ttfamily\fontsize{6}{7}\selectfont, text=black!70, anchor=north, inner sep=2pt] at (\x,-0.05) {lvl \lab};
}
\node[font=\sffamily\tiny, text=black!55, anchor=west, inner sep=2pt] at (12.6,0.0) {harder};
\end{tikzpicture}% 
};
% --- 4. Forking Arrows ---
\draw[arrow, rounded corners=4pt] (api.south) -- +(0,-2.5mm) -| ([yshift=1pt]p1.north);
\draw[arrow, rounded corners=4pt] (api.south) -- +(0,-2.5mm) -| ([yshift=1pt]p2.north);
% --- 5. Frame wrapping all content tightly ---
\node[draw=black!12, rounded corners=10pt, line width=0.6pt, inner sep=6pt,
fit=(api)(p1)(p2)(diff)] {};
\end{tikzpicture}
\caption{\textbf{System overview.} Reasoning Core exposes a unified task API for procedurally generating verifiable symbolic reasoning data. The same calls return training-ready examples (\texttt{prompt}, \texttt{answer}, optional trace) for symbolic pre-training / SFT, and support RLVR via algorithmic verification (\texttt{score\_answer}). Task examples are illustrative.}
\label{fig:interface}
\end{figure*}
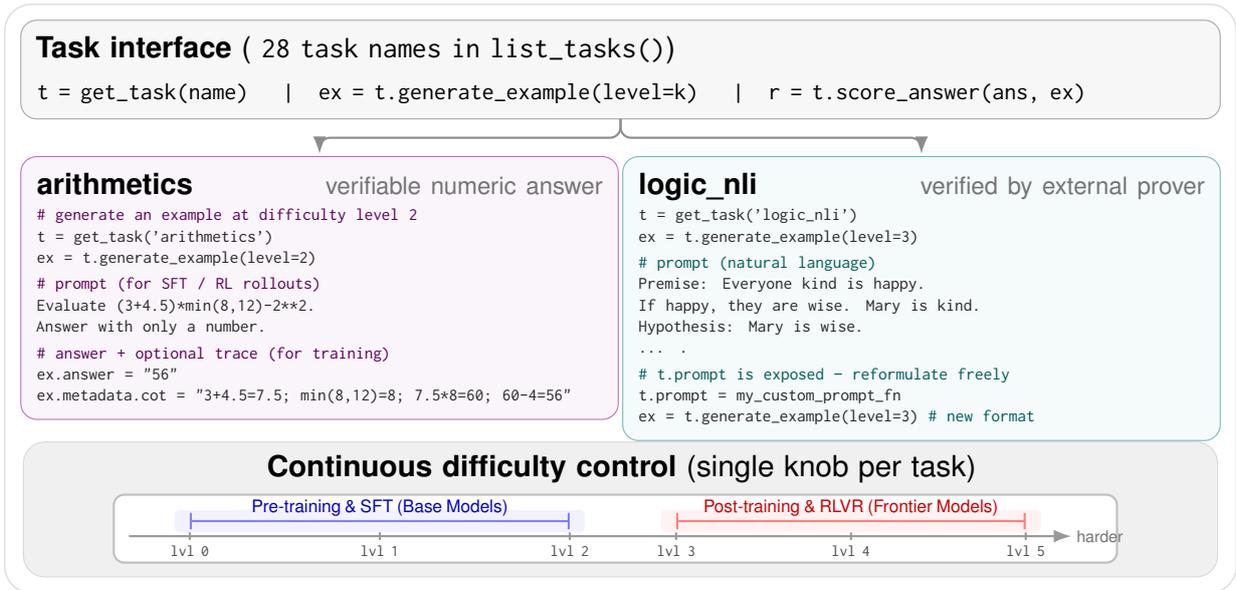

\paragraph{Pre-training on procedurally generated data}
Recent studies have shown that procedurally generated formal data during pre-training can instill foundational reasoning primitives and useful inductive biases~\citep{hu-etal-2025-circuits, jiang2026proceduralpretrainingwarminglanguage, Wu2022InsightsIP, allen2023physics}. These efforts have largely relied on simple formal languages, most prominently Dyck grammars for modeling hierarchical dependencies. Reasoning Core extends this line of work with substantially broader domain coverage (full context-free grammars, including Dyck, alongside our other tasks), while remaining compatible with post-training via its verifiable reward interface. To our knowledge, it is the first library to equip symbolic pre-training data with external solver verifiers for reliable correctness guarantees and to include rich reasoning traces that can seed chain-of-thought behaviors from the earliest training stages.

\section{Reasoning Core \label{sec:rc_features}}

\begin{figure*}[ht]
\centering
\includegraphics[width=\textwidth]{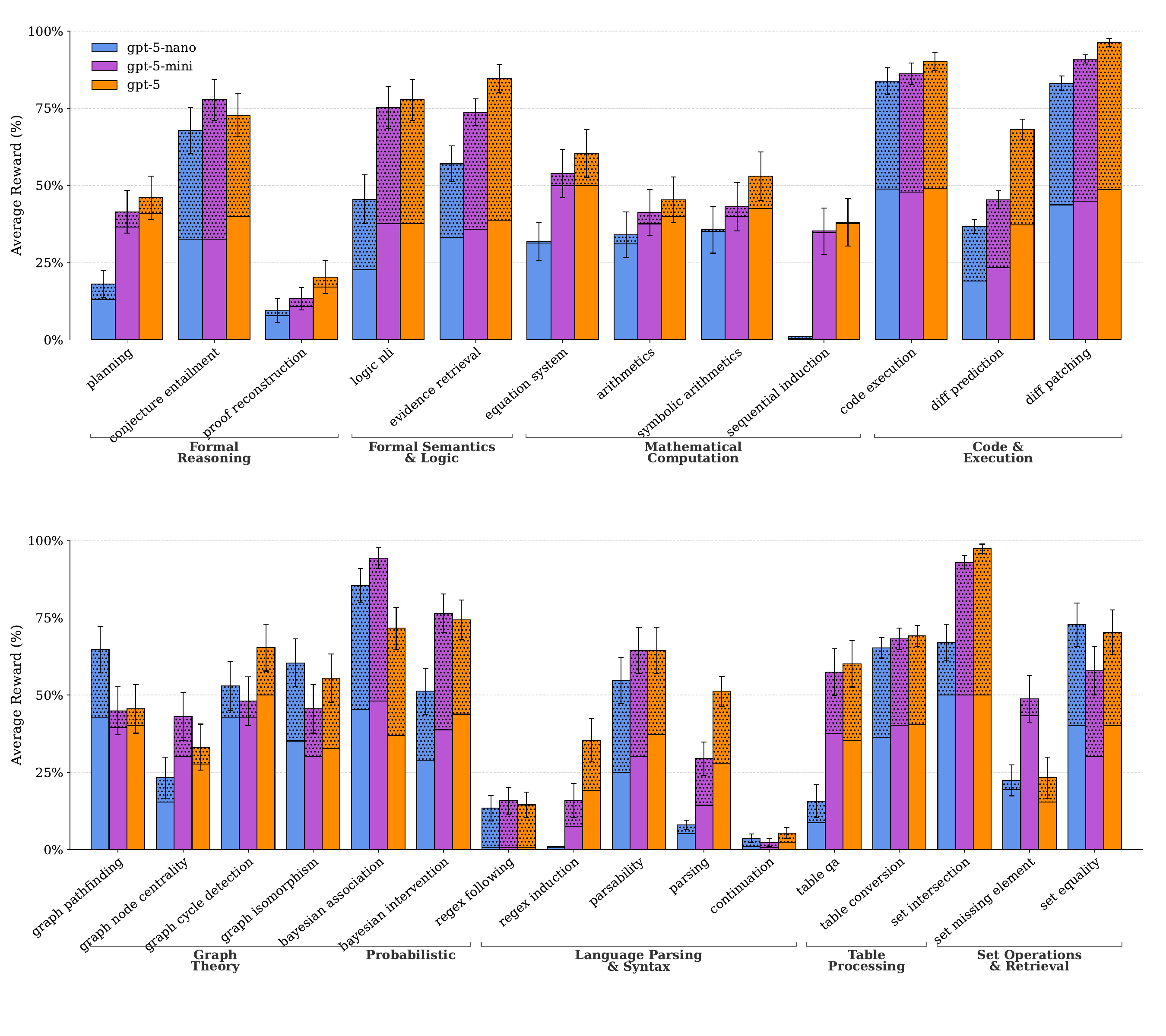}
\caption{Zero-shot average reward of GPT-5 on Reasoning Core tasks across two difficulty levels (\easySymbol\ easy, \hardSymbol\ hard), with standard error. All tasks are challenging, especially at higher difficulty.}
\label{fig:tasks}
\end{figure*}

\begin{figure*}[ht]
\centering
\includegraphics[width=0.9\textwidth]{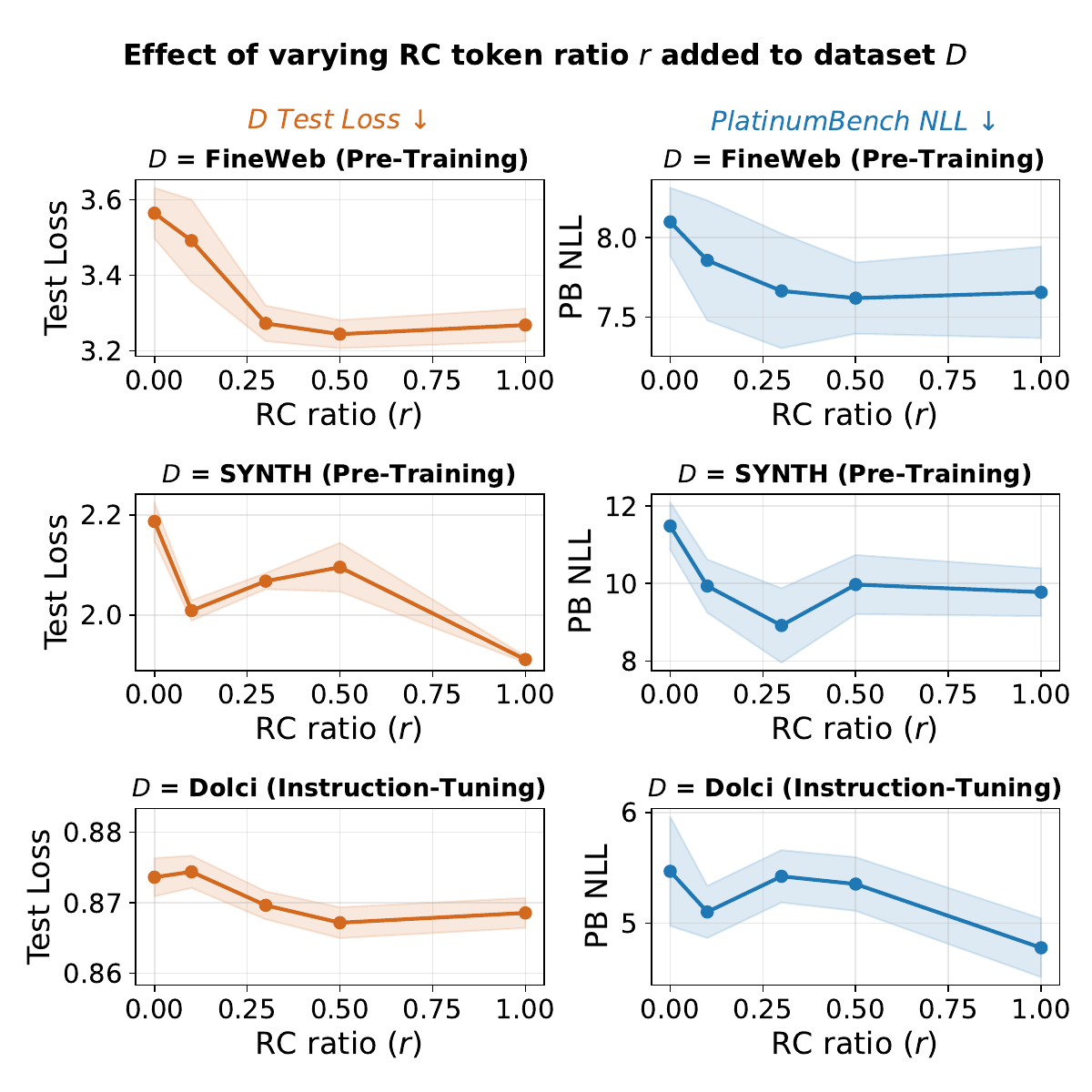}
\caption{
Test NLL on various training data $D$ and test answer NLL on PlatinumBench reasoning tasks over two runs, with standard error.
}
\label{fig:sft}
\end{figure*}

Reasoning Core is designed around four properties: generality, readability, scalability, and rigorous verification.

\subsection{High Generality and Foundational Tasks}
Our task selection targets fundamental formal reasoning capacities rather than narrow puzzles. It includes PDDL planning over randomized domains, first-order logic with equality, non-linear equation systems, context-free grammar parsing and generation, regex and symbolic induction, lightweight retrieval/perception tasks (e.g., missing-element detection, set intersection), causal reasoning over random Bayesian networks, and formal mathematics tasks via the TPTP ecosystem (e.g., axiom/theorem matching, useful-axiom detection, proof structure).

Task descriptions are available in Appendix \ref{sec:task_desc}.

\subsection{Scalable Generation and Difficulty Control}
Each generator is controlled by a continuous ``difficulty knob'', a single float that parametrically adjusts underlying factors (e.g., proof depth, number of variables, plan length). Inherently discrete hyperparameters use stochastic rounding, enabling fine-grained control over the problem distribution. This design supports adaptive curricula tailored to a model's evolving performance.

\subsection{Verification via External Tools}
Internal verification alone is hard to scale to wide problem distributions without reinventing a solver. We therefore integrate external tools such as theorem provers for logic (Vampire/E), planning engines for PDDL (FastDownward), and symbolic algebra systems for equation solving (Sympy) to provide objective and unambiguous reward signals. Solvers with hard non-Python dependencies are containerized via udocker/Apptainer with custom wrappers we provide.

\subsection{Grammar-Based Generation \label{sec:gram}}
We release \texttt{gramforge}\footnote{https://github.com/sileod/gramforge}, a framework originally introduced for logical reasoning datasets \citep{sileo-2024-scaling}. The library allows the definition of context-sensitive probabilistic grammars where production rules are synchronized across multiple output channels (e.g., generating natural language and its corresponding logical form simultaneously). Separating the generation algorithm from the grammar allows concise, auditable, and expressive generators. We introduce several key enhancements to the generation algorithms and available resources:

\paragraph{Topological Control} Standard PCFG sampling often struggles to satisfy minimum depth constraints without degenerating into ``spikes,'' deep but narrow derivation trees that lack syntactic diversity. We implement a custom generation algorithm with a \textit{bushiness} factor. This parameter forces lateral expansion alongside vertical growth, ensuring that generated examples maintain structural complexity rather than just length.

\paragraph{Context Sensitivity} To support procedural code generation, the library now handles state propagation during derivation. This allows for consistent tracking of variable scopes and loop invariants, which are impossible in pure Context-Free Grammars.

\paragraph{Pre-loaded Grammars} We provide a suite of implemented grammars. We reimplement \textit{TinyPy}, a grammar for Python code synthesis \citep{yamani2024automatic} for our code execution task, and propose a simplified English grammar for our syntax tasks, alongside the English/TPTP First-Order Logic grammars used in our logic tasks, a regex grammar, and a CFG meta-grammar.

\subsection{Efficient Data Production}

Many Reasoning Core tasks rely on rejection sampling and external solvers with highly variable runtimes. To enable robust generation at scale, timeouts are automatically scaled with difficulty level, and stalled external processes are detected and killed to avoid resource leaks. A \emph{balancing key} mechanism caps the frequency of task-declared features (typically answer labels) within each batch, preventing degenerate distributions. We also provide a parallel generation pipeline where single-threaded workers coordinate via file locks, scaling linearly across CPU cores.

\subsection{Reasoning traces \label{sec:cot}}
When possible, we add reasoning traces to the examples. They can be used to seed pseudo-chain-of-thought behaviors. For example, we re-format TPTP proofs in the logic and mathematical tasks. When this is not possible (e.g., the arithmetic tasks), derivations are automatically generated by logging intermediate steps during a recursive, bottom-up evaluation of the expression tree using exact fractional arithmetic.

\section{Experiments}

We generate datasets by randomly sampling tasks with the unified interface and parallel generation utilities, along with timeouts and safeguards. For the pre-training dataset, 80\% of examples use difficulty level 0; the remaining 20\% is split between levels 1 and 2. For the post-training dataset, we sample the difficulty uniformly among 0, 3 and 5. We release 10M examples (5B tokens) for pre-training and 1M examples (1B tokens) for post-training. Generating each subset took about 3 days with 48 threads on Intel(R) Xeon(R) Gold 5320.

\subsection{Zero-shot evaluation}
We conducted an initial zero-shot evaluation of the GPT-5 family on the Reasoning Core tasks, assessing performance across two difficulty levels: easy (knob level 0) and hard (knob level 5), with 200 samples per task and difficulty. We set the \texttt{reasoning effort} to \texttt{medium} and used the default hyperparameters (temperature 1, top-p 1), with \textit{Output the answer in the desired format between <answer> and </answer>} as a system prompt. Figure \ref{fig:tasks} presents the average reward obtained for each task configuration. All tasks are sufficiently challenging for GPT-5, and the difficulty control works as intended for most tasks, leading to higher failure rates in hard mode.
While this experiment does not substitute for an RLVR evaluation, it shows that the datasets we generate are challenging even for large frontier models, for both pre-training and post-training.

\subsection{Supervised fine-tuning}

We use the TRL library \cite{vonwerra2020trl} to perform supervised fine-tuning experiments on three datasets: FineWeb \cite{penedo2024fineweb} and SYNTH \cite{pleias2025synth} for pre-training, and Dolci \cite{olmo2025olmo} for instruction-tuning (post-training).
For pre-training, we use a randomly initialized Transformer with the Monad-56M \cite{pleias2025synth} architecture.
For instruction-tuning, we use the pre-trained Ettin-68M decoder \cite{ettin2025}.
Model and data sizes are chosen to approximate Chinchilla-optimal ratios~\cite{hoffmann2022training}; each run takes roughly one day on a single Nvidia A30 GPU.
We use the Prodigy \cite{mishchenko2024prodigy} optimizer with the Schedule-Free algorithm \cite{schedulefree} and default hyperparameters, a batch size of 16, and a context window of 1024 tokens, over one epoch.

In all cases, we use 0.5B tokens of the natural language datasets $D\in \{\text{FineWeb, Dolci, SYNTH\}} $ to which we add $r\times $0.5B tokens from Reasoning Core, for $r \in \{0, 0.1, 0.3, 0.5, 1.0\}$. When available, we include the reasoning traces (Section~\ref{sec:cot}) for $50\%$ of the Reasoning Core examples.

We evaluate on both the test set of each dataset and on PlatinumBench \cite{platinumbench}, which contains 15 tasks targeting reasoning reliability across domains (math, logic, table understanding).
Negative Log Likelihood (NLL) is the standard language modeling loss on answers, and is a continuous metric that is less noisy than accuracy because small models can have trouble following instructions and obtaining meaningful accuracy results.

Results in Figure \ref{fig:sft} show that combining the datasets consistently improves PlatinumBench answer NLL across all three corpora. It also slightly reduces the validation loss on the general natural language training data, confirming that symbolic reasoning primitives aid natural language modeling.

The sweet spot seems to be adding half as many training tokens as the original size (r=0.5) to the given corpora, leading to a third of symbolic tokens. Mixing does increase the total token count, but procedurally generated data can be produced at negligible marginal cost, unlike finite web corpora. The key finding is that these additional tokens improve reasoning without degrading language modeling.

While this experiment confirms the difficulty of our tasks, we do not report RLVR training results in this work. Although all tasks natively expose verifiable reward functions, the contribution of this paper is the generation infrastructure and its integration with external solvers; a meaningful RLVR study at the scale required to master these high-generality distributions demands substantially larger compute budgets and careful multi-task balancing. We scope large-scale RL validation to future work.

\section{Conclusion}
Training on verifiable symbolic data is a promising path toward building, rather than merely surfacing, reasoning capabilities in language models, and wider data distributions become increasingly necessary as training scales grow. We introduced Reasoning Core to support this direction: a scalable data generation suite that targets fundamental symbolic domains with high distributional generality, paired with external solver verification and continuous difficulty control. Our experiments demonstrate that mixing Reasoning Core data into both pre-training and instruction-tuning consistently improves downstream reasoning while preserving, or slightly improving, language modeling quality, and zero-shot evaluations confirm that these tasks challenge even frontier models such as GPT-5. Because the same interface exposes verifiable reward functions, Reasoning Core is also suited for reinforcement learning; we leave this direction, along with investigations of transfer to non-symbolic domains, to future work.

\section*{Limitations}

\paragraph{Scope} The scope of Reasoning Core is inherently circumscribed to formal and symbolic domains (planning, logic, grammar parsing, equation solving, and related tasks). While the reasoning capacities exercised by these domains (e.g., systematic search, compositional inference, symbolic manipulation, planning) are plausibly transferable to less formal settings such as legal reasoning, scientific hypothesis evaluation, ordering agentic tool calls, or structured argumentation, we have not empirically validated such transfer. Investigating cross-domain generalization remains important future work.

\paragraph{Scale} Our experiments are also conducted at limited scale: fine-tuning runs use models under 100M parameters trained on 0.5B tokens, and zero-shot evaluations, while informative, do not constitute a full training study. Whether the benefits and trade-offs we observe hold at larger model and data scales is an open question.

\paragraph{Absence of RLVR Experiments}
Although Reasoning Core is suitable for Reinforcement Learning with Verifiable Rewards (RLVR), we do not present RLVR training curves in this paper. In our setting, the generators intentionally target high distributional breadth (e.g., random PDDL domains, full first-order logic) to prevent overfitting. Consequently, learning these tasks is substantially more sample-intensive than mastering templated or low-entropy procedural datasets. As a result, small-budget RLVR experiments (e.g., a few thousand episodes) are unlikely to be representative of the regime in which these environments are intended to be used and could underestimate their utility. A rigorous RLVR evaluation requires massive rollouts, careful multi-task mixing, and sophisticated curriculum balancing strategies, which we leave to future work.

Finally, although we invested substantial effort in verifying the correctness of generated problems, combining external solver checks (theorem provers, planning engines, symbolic algebra systems) with LLM-assisted review and multiple rounds of independent human adjudication and code auditing, no verification pipeline is infallible. Subtle bugs in generator logic, edge cases missed by solvers, or mismatches between a problem's natural-language presentation and its formal specification may persist. Users should be aware that a small fraction of generated instances could contain errors despite these safeguards.

\section*{Broader Impact}
The ability to generate virtually unlimited verifiable symbolic data opens largely unexplored avenues for language model pre-training. While post-training optimization is well established, injecting procedural formal structures at the pre-training or mid-training stages offers a scalable path toward neurosymbolic AI. By internalizing logical rules and structural constraints directly into their weights \citep{Clark2020TransformersAS}, models can develop robust foundational reasoning capabilities before being exposed to standard natural language corpora. Reasoning Core provides the necessary infrastructure to test and scale this approach across diverse formal domains.
Furthermore, fully procedural data circumvents the severe licensing ambiguities and copyright risks associated with web-scraped datasets or text distilled from proprietary language models \citep{longpre2024large}. Because every instance is procedurally novel, Reasoning Core is also immune to benchmark contamination, a growing concern for static evaluation suites. Furthermore, the data we generate can be re-validated with our code as a safeguard against data poisoning \cite{bowen2025scaling,cloud2025subliminal} providing a verifiable pre-training component. Prompts are perfectly reconstructible from explicitly disentangled metadata, and solution correctness can be deterministically re-evaluated at any time using the provided scoring interfaces. By offloading complex verification to established external solvers, the generation codebase remains concise and easily inspectable. This design ensures that the resulting datasets are legally safe and reproducible.

\section*{Acknowledgements}
This work was supported by the French National Research Agency (ANR) through the ANR-24-CE23-4637 grant (Adada project).

\bibliographystyle{acl_natbib}
\bibliography{custom}
\clearpage
\appendix
\section{Task descriptions \footnote{See examples in \href{https://github.com/sileod/reasoning_core/blob/main/GALLERY.md}{GALLERY.md} not included due to appendix size limit.} \label{sec:task_desc}}

\newcommand{\taskdesc}[2]{\noindent\textbf{\texttt{#1}}\quad #2\par\vspace{2pt}}

\taskdesc{arithmetics}{Evaluate arithmetic expressions (addition, subtraction, multiplication, division, exponentiation, \texttt{abs}, \texttt{min}, \texttt{max}, \texttt{round}) generated by sampling random expression trees up to a configurable depth, with integer or float leaves. Difficulty scales with tree depth and output precision. Chain-of-thought traces evaluate bottom-up, one step per line.}

\taskdesc{symbolic\_arithmetics}{Simplify algebraic expressions over symbolic variables and integer constants, generated by a probabilistic context-free grammar over arithmetic operations and functions. Variable count, numeric range, and expression depth scale with difficulty. Answers are canonicalized via symbolic simplification; scoring is equivalence-based, tolerating algebraically identical reformulations.}

\taskdesc{equation\_system}{Solve a system of linear equations for a queried variable. Systems may be uniquely determined, underdetermined (\textit{Multiple solutions}), or inconsistent (\textit{No solution}), each case sampled probabilistically. Unique systems are constructed via random integer solutions with coefficient-mixing row operations of variable complexity.}

\taskdesc{conjecture\_entailment}{Decide whether premises entail a conjecture under the superposition calculus, following \citet{quesnel2025saturation}. Problems are extracted from saturation of TPTP axioms across diverse mathematical domains at controlled proof depths. Negative examples are produced by premise perturbation and verified by an automated theorem prover.}

\taskdesc{proof\_reconstruction}{Reconstruct the dependency graph of a formal proof: given shuffled, numbered clauses, identify axioms vs.\ derived clauses and output parent--child edges (\texttt{CHILD <- P1, P2}). Subgraphs are extracted from theorem-prover runs over TPTP axiom sets across ten mathematical domains, with a strict binary-inference constraint. Scored by F1 over derivation edges.}

\taskdesc{logic\_nli}{Natural language inference over verbalized first-order logic formulae, following \citet{sileo-2024-scaling}. Premises are sampled FOL sentences over randomized vocabularies of names and adjectives; labels (entailment, contradiction, neutral) are verified by the Vampire theorem prover. The distribution spans arbitrary quantifier, negation, and predicate combinations.}

\taskdesc{evidence\_retrieval}{Extending \texttt{logic\_nli} following \citet{sileo2025logic}: given numbered premises and a hypothesis, identify the minimal premise subset that entails or contradicts the hypothesis. Gold answers are verified by ablation---a premise is included only if removing it invalidates the proof.}

\taskdesc{planning}{Produce a valid sequential plan for procedurally generated STRIPS-style planning problems presented in structured natural language. Both domains (actions, preconditions, effects) and instances (objects, initial state, goal) are randomly generated. Plans are deterministic via lexicographic action ordering. Chain-of-thought traces track state changes and remaining goals. Scoring function accepts all valid plans with length penalty.}

\taskdesc{graph\_pathfinding}{Find the shortest path between two nodes in an undirected graph sampled from diverse topologies (Erd\H{o}s--R\'enyi, Watts--Strogatz, Barab\'asi--Albert, regular, grid), presented in randomized text formats (edge lists, adjacency lists, matrices, DOT, etc.). Chain-of-thought traces follow BFS execution, tracking queue state and visited sets.}

\taskdesc{graph\_node\_centrality}{Identify all nodes with highest degree centrality in an undirected graph sampled from diverse topologies and rendered in varied text formats. Output is a sorted list of all maximally central nodes, with ties handled explicitly.}

\taskdesc{graph\_cycle\_detection}{Identify the exact node set forming the unique cycle in a graph constructed by adding one edge to an acyclic path backbone. Graphs use randomized text representations. Output is a sorted node list; difficulty scales with graph size.}

\taskdesc{graph\_isomorphism}{Determine whether two graphs are isomorphic. Graphs are sampled from diverse topology generators with independently chosen text renderings. Positive pairs use random node permutations; negatives use edge swaps, confirmed non-isomorphic. The class distribution skews toward negative examples.}

\taskdesc{sequential\_induction}{Infer a recurrence relation $U[n]=f(U[n\!-\!1],\ldots,U[n\!-\!d],n)$ from an integer sequence prefix. Expressions are sampled from a context-free grammar over arithmetic operators and prior terms, filtered for non-degeneracy (no constant or exploding sequences). Scoring rewards both correctness on observed terms and formula conciseness.}

\taskdesc{bayesian\_association}{Compute the posterior marginal of a target variable in a random Bayesian network given observational evidence (Pearl's Rung~1). Networks have discrete variables with tabular or noisy-interaction CPDs (e.g., Noisy-AND/OR). Scored via Jensen--Shannon divergence against ground truth.}

\taskdesc{bayesian\_intervention}{Compute the posterior under a \textit{do}-intervention (Pearl's Rung~2) in a random Bayesian network. The model must apply graph surgery---severing incoming edges to the intervened variable---then propagate the resulting distribution. Same network distribution and JSD scoring as \texttt{bayesian\_association}.}

\taskdesc{code\_execution}{Predict the standard output of a synthetic Python program sampled from a hierarchical grammar (\texttt{tinypy}) spanning simple arithmetic assignments to richer control flow and expressions. A continuous difficulty parameter controls the distribution over construct complexity levels.}

\taskdesc{diff\_prediction}{Given two versions from a stochastically mutated plain-text file history (line insertions, deletions, word substitutions), produce the correct unified diff in chunk format. Source may be older or newer than target; the diff may be empty. Scored by character-level similarity.}

\taskdesc{diff\_patching}{Apply a unified diff patch to a source text to produce the target text. Documents evolve through stochastic mutations; patches range from empty to multi-hunk edits, ensuring coverage of trivial and complex patching scenarios. Scored by character-level similarity.}

\taskdesc{parsability}{Given a context-free grammar and a token string, classify as \textit{unambiguous}, \textit{ambiguous}, or \textit{unparsable}. Grammars are randomly generated via a meta-grammar or drawn from curated sets (e.g., Dyck languages); strings may be perturbed for negative examples. Labels are determined by Earley parse counts; chain-of-thought traces each parse path.}

\taskdesc{parsing}{Produce a fully parenthesized Lisp-style parse tree for a token sequence under a given CFG. Grammars are sampled or procedurally generated; only unambiguous parses are retained. A \emph{tagging} variant requires labeling each token with its parent nonterminal and parse-tree depth.}

\taskdesc{continuation}{Identify all valid next tokens after a prefix string under a given CFG, plus whether the prefix is already a complete sentence (\texttt{STOP}). Valid continuations are computed exactly via Earley parsing over diverse sampled and curated grammars. Scored by token-set overlap with partial credit.}

\taskdesc{regex\_following}{Produce a string that fully matches a given regular expression sampled from a stochastic grammar over character classes, quantifiers, alternation, grouping, and ranges. Reward is inversely proportional to the fuzzy edit distance between the model's output and the target pattern.}

\taskdesc{regex\_induction}{Infer a regular expression matching all positive examples while rejecting all negatives. Regexes are sampled from a probabilistic grammar of varying complexity; positives and negatives are verified against the target. Scoring rewards correctness (match/reject rates) and conciseness.}

\taskdesc{set\_intersection}{Compute the intersection of two sets from a random domain (integers, English/French number words, dates, alphabetic strings). Overlap between sets is controlled to ensure non-trivial instances. Scored via Jaccard similarity, rewarding partial credit.}

\taskdesc{set\_missing\_element}{Identify 1--3 elements removed from a contiguous subsequence of an ordered domain (integers, number words, dates, alphabetic strings), presented in shuffled order, following \citet{sileo2024attention}. Difficulty scales with sequence length and domain variety.}

\taskdesc{set\_equality}{Determine whether two shuffled sets contain exactly the same elements (\texttt{True}/\texttt{False}). The second set is either an exact shuffled copy or perturbed by adding, removing, or replacing elements. Domains span integers, number words, dates, and alphabetic strings.}

\taskdesc{table\_qa}{Execute a SQL query over a synthetic table with semantically typed columns (prices, ratings, names, cities, etc.), rendered in a random format (CSV, Markdown, JSON, \LaTeX, YAML, etc.). Queries span aggregations, filters, grouped summaries, and distinct counts. Scoring handles numeric tolerance and structured comparison.}

\taskdesc{table\_conversion}{Convert a table between format pairs drawn from CSV, JSON, YAML, Markdown, HTML, \LaTeX, and plain text. Tables have varied semantic column types; source and target formats are sampled uniformly from all pairwise combinations. Scored by normalized edit distance.}

\end{document}